\documentclass[lettersize,journal]{IEEEtran}
\usepackage{amsmath,amsfonts}
\usepackage{algorithmic}
\usepackage{algorithm}
\usepackage{array}
\usepackage[caption=false,font=normalsize,labelfont=sf,textfont=sf]{subfig}
\usepackage{textcomp}
\usepackage{stfloats}
\usepackage{url}
\usepackage{verbatim}
\usepackage{graphicx}
\usepackage{cite}

\usepackage{graphicx}
\usepackage{multirow}
\usepackage{stmaryrd}
\usepackage{diagbox}
\usepackage{amsmath}

\begin{document}

\title{Spatial-temporal traffic modeling with a fusion graph reconstructed by tensor decomposition}

% \author{IEEE Publication Technology,~\IEEEmembership{Staff,~IEEE,}
%         % <-this % stops a space
\author{Qin Li, Xuan Yang, Yong Wang, Yuankai Wu, Deqiang He~\IEEEmembership{}

% \thanks{This paper was produced by the IEEE Publication Technology Group. They are in Piscataway, NJ.}% <-this % stops a space
% \thanks{Manuscript received April 19, 2021; revised August 16, 2021.}
\thanks{Manuscript received November 26, 2022; The work was supported by NSFC (U22A2053).\textit{(Corresponding author: Qin Li)}}
\thanks{Qin Li, Xuan Yang, and Deqiang He are with the School of Mechanical Engineering, Guangxi University, No.100 East Daxue Road, Xixiangtang District, Nanning Guangxi 530004, China,( E-mail:lqfy123@gxu.edu.cn) }

%\thanks{Xuan Yang: School of Mechanical Engineering, Guangxi University, No.100 East Daxue Road, Xixiangtang District, Nanning Guangxi 530004, China (E-mail:y$\_$xuan0714@163.com)}

\thanks{Yong Wang is with the School of Mechanical Engineering, Beijing Institute of Technology, No.5 Yard, Zhong Guan Cun South Street, Haidian District, Beijing 100081, China} %(E-mail:17862709675@163.com)}

\thanks{Yuankai Wu is with the College of Computer Science, Sichuan University, Wangjiang road 29, Wuhou District, Chengdu Sichuan 610064, China}%(E-mail:wuyk0@scu.edu.cn)}

%\thanks{Deqiang He: School of Mechanical Engineering, Guangxi University, No.100 East Daxue Road, Xixiangtang District, Nanning Guangxi 530004, China (E-mail:hdqlqy@gxu.edu.cn)}
}
% The paper headers
% \markboth{Journal of \LaTeX\ Class Files,~Vol.~14, No.~8, August~2021}%

%% 
% \markboth{ IEEE TRANSACTIONS ON INTELLIGENT TRANSPORTATION SYSTEMS}%
% {Shell \MakeLowercase{\textit{et al.}}: A Sample Article Using IEEEtran.cls for IEEE Journals}

% \IEEEpubid{0000--0000/00\$00.00~\copyright~2021 IEEE}
% Remember, if you use this you must call \IEEEpubidadjcol in the second
% column for its text to clear the IEEEpubid mark.

\maketitle

\begin{abstract}
Accurate spatial-temporal traffic flow forecasting is essential for helping traffic managers to take control measures and drivers to choose the optimal travel routes. Recently, graph convolutional networks (GCNs) have been widely used in traffic flow prediction owing to their powerful ability to capture spatial-temporal dependencies.  The design of the spatial-temporal graph adjacency matrix is a key to the success of GCNs, and it is still an open question. This paper proposes reconstructing the binary adjacency matrix via tensor decomposition, and a traffic flow forecasting method is proposed.   First, we reformulate the spatial-temporal fusion graph adjacency matrix into a three-way adjacency tensor. Then, we reconstructed the adjacency tensor via Tucker decomposition, wherein more informative and global spatial-temporal dependencies are encoded. Finally, a Spatial-temporal Synchronous Graph Convolutional module for localized spatial-temporal correlations learning and a Dilated Convolution module for global correlations learning are assembled to aggregate and learn the comprehensive spatial-temporal dependencies of the road network. Experimental results on four open-access datasets demonstrate that the proposed model outperforms state-of-the-art approaches in terms of the prediction performance and computational cost.
\end{abstract}


\section{Introduction}
\IEEEPARstart{W}{ith} the rapid development of economy and the rising number of vehicles, numerous worldwide cities are gradually plagued by traffic congestion and parking difficulties. Many countries have been striving to develop the intelligent transportation system (ITS) to increase the traffic management efficiency, enhance the travel security and improve the increasingly serious traffic conditions \cite{zhang2011data}. Traffic forecasting is regarded as an important part of the intelligent transportation system, and its researches have gained continuous attention during the past dedicates \cite{zhao2019t,zhang2020spatio}. Generally, traffic flow forecasting methods are roughly classified as the statistical models like the historical average (HA) \cite{smith1997traffic} and the autoregressive integrated moving average (ARIMA) \cite{lippi2013short}, and the machine learning based models \cite{van2012short}. In the past decades, owing to the increasement on the quantity of the traffic flow, its forecasting evolves into a large-scale network traffic forecasting problem which is challenging for statistical models since it is easy for them to fall into dimensionality disasters \cite{karlaftis2011statistical}. Hence, machine learning models, e.g. support vector regression (SVR) \cite{wu2004travel}, gradient boosting regression tree (GBRT) \cite{zhan2018multi} and artificial neural networks (ANN) \cite{huang2014deep}, based traffic flow forecasting methods become the research focus.

Although the above traditional machine learning methods improved the traffic forecasting accuracy to some extent, they are still poor at capturing and analyzing the complex nonlinear correlations of the large-scale network traffic data \cite{cui2019traffic}. Fortunately, deep learning methods, emerging in recent years, make up for this deficiency through an end-to-end multi-layer learning framework. Early deep learning models applied to traffic flow prediction are the recurrent neural networks (RNN) and its variants, such as long short-term memory (LSTM) \cite{hochreiter1997long} and gated recurrent unit (GRU) \cite{zhang2018combining} networks, since they can capture and utilize the temporal correlations of traffic data fully to improve the forecasting accuracy. Subsequently, some researches tried to employ convolutional neural networks (CNN) \cite{zhang2017deep} to capture spatial correlations among the traffic network to further enhance the forecasting accuracy. But CNN works in the Euclidean space, which depicts the correlations of the data through Euclidean distances. However, the structure of the real-world large-scale traffic network is topological and irregular, whose spatial correlations cannot be fully characterised by CNN.

Recently, the graph convolutional neural network (GCN) \cite{yu20193d}, which can characterize and utilize the spatial topological structure of road networks, has been widely used in traffic flow forecasting \cite{yu2017spatio} and has achieved superior performances \cite{zhao2019t,kong2022adaptive,zhang2021traffic}. Existing studies show that constructing the proper adjacency graph instead of adopting sophisticated mechanisms for modeling spatial and temporal correlations is a key to enhancing the graph convolutional networks, and many research efforts have been made. Early efforts mainly concentrated on improving the given spatial adjacency matrix, e.g., \cite{9098104} changed the 0-1 adjacency matrix into  a distance-based weighted adjacency matrix whose entries are usually correlated to the given physical spatial distances between adjacent nodes. However, only utilizing the given spatial adjacency matrix to model the graph is not informative enough, since limited representations of the given spatial matrix fail to depict the temporal similarity between nodes to obtain the complete adjacent connections. Some works already made several attempts to improve the temporal representation of the graph. Self-adaptive matrices \cite{wu2019graph, bai2020adaptive} are employed to adjust the existed spatial adjacency matrix and learn the hidden spatial dependencies, but these learnable matrices are lack of correlations representation ability for complicated spatial-temporal dependencies and the acquisition for them is time consuming. Recently, Song \cite{song2020spatial} introduced a localized spatial-temporal graph construction method through connecting the nodes at different time steps into a graph and incorporating it with the given spatial graph into a fusion graph, and Li \cite{li2021spatial} improved this fusion graph model by deploying the Dynamic Time Warping algorithm to measure the similarity between time series and fusing three types of sub-graphs including the given spatial graph,the temporal graph, and the temporal connectivity graph. Both of them captured the spatial-temporal correlations directly trough a fusion graph model instead of using different types of neural network modules separately, hence made the learning model with less computational cost and more efficient. However, they still need a mask matrix learning process to update the graph adjacency matrix during prediction, which increased the model learning parameters to some extent. Besides, they only constructed the graph through a block matrix with each block representing a sub-graph and update the weights of entries in each block separately, which ignored the possible potential temporal-spatial correlations among the blocks.

This paper aims to offline reconstruct the fusion graph through a information maximization mapping method that can  extract the  potential temporal-spatial correlations among the blocks of the original fusion graph to provide a more informative graph and meanwhile avoid the time-consuming online learning process on the graph weights updating. Since the fusion graph is a block matrix that is usually naturally expressed as a tensor model\cite{cichocki2016tensor}.We attempt to deploy tensor decomposition, which has been proved to be capable of extracting the multi-mode spatial-temporal correlations of traffic data efficiently and widely used in  traffic data modeling\cite{chen2018spatial,tan2016short}, to reconstruct the graph and enhance the graph convolutional neural network based traffic data forecasting model. The main contributions of our work are as follows:
\begin{itemize}[]
\item We present a novel tensor decomposition based spatial-temporal graph reconstruction method that can maximally utilize the  spatial-temporal information  and extract the possible potential correlations among the sub-graphs to offline reconstruct and provide a more informative fusion graph that helps to contribute to the prediction accuracy and yet the computational cost.
\item We assemble the spatial-temporal graph convolutional module (STTGCM) \cite{song2020spatial} and the Dilated Convolution module into a new spatial-temporal graph convolutional network model (STTGCN) to learn the localized and global spatial-temporal correlations simultaneously for traffic data forecasting. 
\item Extensive experiments are conducted on four real-world datasets and the experimental results show that our model consistently outperforms state-of-the-art approaches.
\end{itemize}

\section{Related Works}
\subsection{Graph Convolution Network}
Traditional convolution has a strong ability to explore local spatial correlations, but it is mainly qualified for handling standard grid data rather than non-Euclidean data such as the  rode-network traffic data. Recently, graph convolution networks, which can remedy this limitation and capture the structural characteristics of networks, have been developed rapidly. There are mainly two categories of graph convolution neural networks: Spectral GCN and Spatial GCN. Spectral GCN transforms the convolution networks from the spatial domain into the spectral domain for convenience of leveraging the spectral theory to conduct convolution operation on the graph directly. The SCNN model proposed by \cite{estrach2014spectral} adopted a learnable diagonal matrix to replace the convolution kernel of the spectral domain and realizes the graph convolution operation. However, it is time-consuming to calculate the eigenvalue decomposition of the Laplacian matrix and the model parameters are complex. In order to solve the above problems, ChebNet model was proposed by \cite{defferrard2016convolutional}, which used Chebyshev polynomials instead of convolution kernels in the spectral domain, and saves time without decomposition. To further simplify the model,  \cite{kipf2016semi} proposed the first-order ChebNet model, in which each convolution kernel has only one parameter. Spatial GCN generalizes the traditional convolutional network from the Euclidean space to the vertice domain directly. The GNN model \cite{hechtlinger2017generalization} uses the random walk method to transform the graph structure data, but it tried to force a graph to be a rule structure. GraphSAGE \cite{hamilton2017inductive} sampled the neighbor vertices of each vertex in the graph, and then aggregated the information contained in the neighbor vertices according to the aggregation function. GAT \cite{velickovic2017graph}, a powerful GCN variant,  aggregated the features of neighboring vertices to the center vertex through using attention layers to adjust the importance of neighbor nodes dynamically.

\subsection{Spatial-temporal Traffic flow Forecasting with Deep Learning Models}
Traffic flow forecasting belongs to correlated time series analysis (or multivariate time series analysis) and has been studied for decades. Owing to the strong learning and nonlinear representation abilities, deep learning based traffic flow forecasting methods have been extensively researched and gained outstanding achievements. Most previous studies of this kind \cite{ma2015long,li2018short,chan2021short} have relied on LSTM or GRU to model the temporal dynamics of the sequential traffic  data. Some studies use temporal convolutional networks \cite{zhao2019deep} to enable the model to process extremely long-time sequences with less time. Although employing temporal correlations of traffic data fully to improve the prediction accuracy, these models neglects the spatial correlation among the traffic series from different sources (spaces/regions/sensors) which may have a great influence on the prediction results. Then, CNN that is apt at learning the spatial correlation was introduced to  capture temporal and spatial correlations simultaneously with RNN to improve the prediction accuracy\cite{jiang2019deepurbanevent,wang2018cross,8694956,lv2018lc}. 

 Although the hybrid model of CNN and RNN improved the prediction performances significantly, it can only deal with data in the Euclidean space, hence fail to fully represent and dig the comprehensive structural information of the real-world large-scale road network that is spatially irregular and topological. Subsequently, GCN-based models that process the road network as a graph directly, avoiding to destroy the topological structure are utilized for more practical traffic data forecasting. Some works\cite{bai2020adaptive,8959420} deployed a spectral-domain GCN to capture the spatial correlations among traffic series and used RNN to model and aggregate the temporal correlations to improve the prediction accuracy. Many recent studies \cite{guo2019attention,zheng2020gman} further use complex spatial-temporal attention mechanisms to help a spectral-domain GCN to capture dynamic spatial-temporal correlations. However, the spectral-based method, in which the graph structure is fixed, is computationally complex. Recently, researches gradually pay attention to the Spatial GCN based traffic data forecasting models\cite{yu2017spatio,li2017diffusion,wu2019graph,song2020spatial,li2021spatial}, and demonstrated the main concerns to improve the prediction accuracy are to  construct a adjacency graph as informative as enough and  to implement effective convolution on the graph to aggregate the spatial-temporal features.

\subsection{Deep Learning Models Optimization through Tensor Decomposition}
Tensor decomposition is the higher-order generalization of  matrix decomposition \cite{kolda2009tensor}. It decomposes high-dimensional tensors into a sum of products of lower dimensional factors. Due to its strong capability of uncovering underlying the hidden low-dimensional structure of high-dimensional data, tensor decomposition has been widely used for data compression, dimension reduction, missing data recovery, and
etc \cite{tan2013tensor}. 

One of the features and benefits of deep learning models is the deep structure, which whereas leads to numerous model training parameters and low training efficiency. Motivated by the data compression methods, some researchers tried to utilize tensor decomposition to  compress the neural model for deep learning to make a trade-off between  the precision and computational efficiency. There are mainly three kinds of tensor decomposition based neural model compression methods. The first kind is to compress the whole deep learning architectures through constructing the corresponding tensor network representation, which has been successfully
applied to Restricted Boltzmann Machines and convolutional architectures\cite{cohen2016convolutional}. The second kind leverages tensor decomposition on the single layers of the network. For instance, \cite{calvi2019tucker} introduced the Tucker Tensor Layer as an alternative to the dense weight-matrices of neural networks. Besides, \cite{yang2016deep} introduced a multitask representation learning framework leveraging tensor factorisation to share knowledge across tasks in fully connected and convolutional DNN layers. 

The use of tensor decomposition methods in graph data processing appears a lively research field recently. Tensor  decomposition extracts the hidden information or main components in the original higher-order data. Graph neural networks perform end-to-end calculations on graph data. However, these graph data contain plenty of potential information. Hence, tensor decomposition is adopted to mine the hidden information of the graph data for improving the performances of the graph neural networks.  For instances, to reduce the computational burden, Xu \cite{xu2021spatial} utilized Tucker decomposition to perform separate filtering in small-scale space, time, and feature modes, while Zhao \cite{zhao2022multi} used tensor decomposition to extract the graph structure features of each view in the common feature space.Such kinds of models have also been  adopted to traffic data forecasting\cite{diao2019dynamic,li2020two}.

It can been seen that the cooperation of tensor decomposition helps improve the efficiency and accuracy of deep learning including the graph neural networks models.

\section{Preliminaries}\label{}
\textbf{Definition 1: }This study uses $\mathcal{G}=\left(V, E, A\right)$ to define spatial networks: $|V|=N$ is the vertices set, in which $N$ represents the number of all nodes (each node represents the position of the corresponding sensor), and $E$ is the set of edges, which reflects the connection relationship between adjacent nodes. $A \in R^{N \times N}$ represents the adjacency matrix and stores the connection information, each of which represents the connection between the corresponding sections. $\mathcal{G}$ represents the relationship between nodes in the spatial dimension.

\textbf{Definition 2: }Graph signal matrix $X^{(t)} \in R^{N \times C}$ : $C$ represents the number of features, $t$ represents the time step, and $X^{(t)} \in R^{N \times C}$ represents the observed value of $t$ step in the figure.

Therefore, the time and space dependencies for traffic prediction modeling can be regarded as: learning the mapping function $\mathcal{F}$ based on the road network $\mathcal{G}$ and road network characteristic matrix $X$. We use $T$ to express the length of the historical spatial-temporal network sequence, $T^{\prime}$ represents the length of the target spatial-temporal network sequence to be predicted, which can be expressed as follows: 

\begin{equation}
\left[X^{(t+1)}, \ldots, X^{\left(t+T^{\prime}\right)}\right]=\mathcal{F}\left(\mathcal{G} ;\left(X^{(t-T+1)}, \ldots, X^{(t)}\right)\right)
\end{equation}

\section{Spatial-Temporal Tensor Graph Neural Network}\label{}
\subsection{Construction of the Spatial-Temporal Tensor Graph}\label{}
\begin{figure}[!ht]
  \centering
  \includegraphics[width=0.5\textwidth]{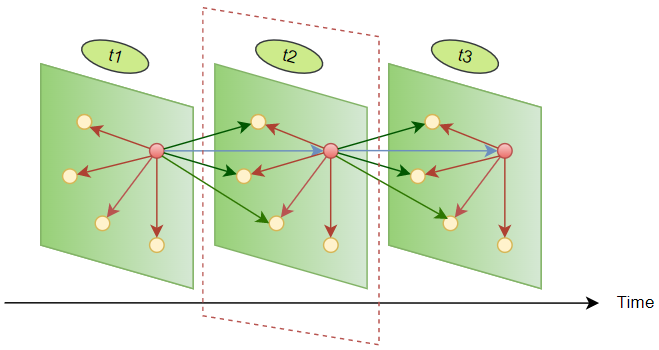}
  \caption{The spatial-temporal influence of the current node in a spatial-temporal network. The red node represents the current node, and the pink arrows represent the impacts of the current node on its neighbors in the $t_{1}$ time slice; The blue arrow indicates the effect of the current node on itself at the next time step, and the green arrows  denote the impacts of the current node on its neighbors at the next time.}\label{fig:1l}
\end{figure}

% \subsubsection{Graph Construction}\label{}
Inspired by \cite{song2020spatial}, as shown in Fig.~\ref{fig:1l}, we unify the relationship of a node with its spatially neighboring nodes and the relationship of the node with itself in neighboring time steps into a graph as input.

\begin{figure*}[!ht]
  \centering
  \includegraphics[width=0.9\textwidth]{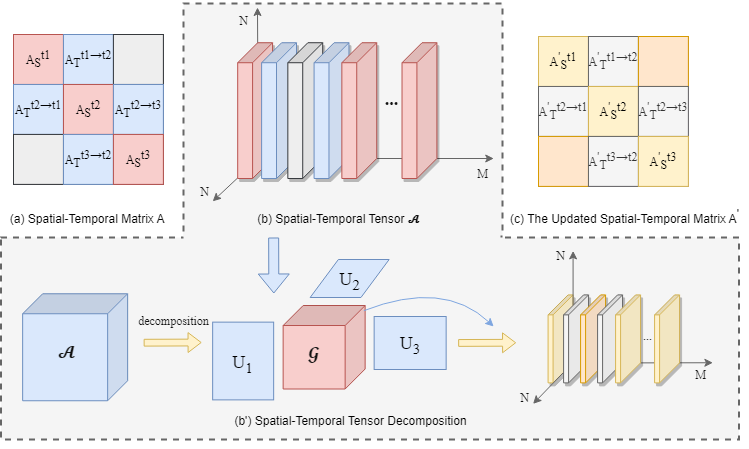}
  \caption{The tensor graph construction. (a) is the spatial-temporal matrix $A$. The pink matrix block represents the impacts of the current node on its neighbors in the $t_{l}(l=1,2,3) $ time step, which belongs to the given spatial connections; The blue matrix block $A_{T}^{t_{l}\rightarrow t_{k}}$  denotes connections between the nodes with themselves at the time step $l$ and $k$. (b) is the spatial-temporal tensor $\mathcal{A}$ constructed by a reshaping operation on the block matrix $A$. $N$ represents the number of nodes and $M$ represents number of blocks in $A$. (b') is the Tucker decomposition  for the spatial-temporal tensor in (b). The core tensor $ \mathcal{G}$ obtained through the decomposition is used to update the graph. (c) is the final graph adjacency matrix ${A}^{\prime}$updated through an unfolding likewise operation on  $ \mathcal{G}$ in (b')}.\label{fig:2l}
\end{figure*}

 As shown in Fig.~\ref{fig:2l}(a), we use $A_{S}^{t_{i}} \in R^{ N \times  N}$ to represent the adjacent matrix
of the given spatial graph in the $t_{i}$ time step and $A_{T}^{t_{l}\rightarrow t_{k}} \in R^{ N \times  N}$ to represent the adjacent matrix of the temporal graph between the adjacent
time steps $t_{l}$ and $t_{k}$, which equals to the connectivity denoted by the blue arrows in Fig.~\ref{fig:1l}.  $A \in R^{ 3 N \times  3 N}$, a  block matrix, is made up of $A_{S}^{t_{i}}$ and $A_{T}^{t_{l}\rightarrow t_{k}}$, which represents the adjacent matrix of the localized spatial-temporal graph constructed from Fig.~\ref{fig:1l}. $A_{S}^{t_{i}}$ or $A_{T}^{t_{l}\rightarrow t_{k}} $ is the single block of  $A$. For each node $i$ in the spatial graph, we recalculate its new index in the local spatial-temporal graph by $(t-1)N+i$, where $t (1 \leq t \leq 3)$ represents the time step number in the local spatial-temporal graph. The entries of $A$ are denoted as:

\begin{equation}
A(i,j)= \begin{cases}  1 ,&\text{if} \quad v_i\quad \text{connects to}\quad v_j \\ 0,&\text{otherwise}\end{cases}
\end{equation}
where $v_i$ denotes the node $i$ in localized spatial-temporal
graph.

However, the 0-1 binary  entries can only depict whether there is connectivity between two adjacent nodes, but fail to show the magnitude of the connectivity (i.e. the weight of each entry). To solve this problem, many existing researches\cite{song2020spatial,li2021spatial}update the weights of each block (sub-graph) separately though learning models, which leads to increment in the volume of model learning parameters and ignores the possible potential correlations among the blocks, e.g, the spatial connectivity in the $t_{1}$ time step $A_{S}^{t_{1}}$ may be correlated with that in the  $t_{2}$ time step $A_{S}^{t_{2}}$. Instead, this paper tries to utilize an information maximization mapping method that attempts to minimize the model parameters and consider the spatial-temporal correlations among the blocks fully to update the weights of all the entries of $A$ simultaneously, i.e. reconstruct the graph. First, for more general and natural spatial-temporal correlations expression \cite{cichocki2016tensor}, we construct the  block matrix $A$ into a tensor model as $\mathcal{A}\in R^{ N \times  N \times M}$ through a reshaping operation,  as shown in Fig.~\ref{fig:2l}(b), in which N represent the number of  nodes in the spatial graph, and $M=9$ represents the number of blocks in $A$. The $(k,l)\textit{th}$ block in the block matrix $A$ is arranged as the $(l+3(k-1))\textit{th}$ lateral slice in the tensor $\mathcal{A}$. Then, Tucker decomposition, a kind of multiple linear mapping, is deployed to reconstruct $\mathcal{A}$. It decomposes $\mathcal{A}$  into a core tensor $\mathcal{G}$ multiplied by the factor matrix $U_{i},i=1,2,3$ along each mode as follows:
 %To solve this problem, instead of updating the weights of each block separately though learning models as many existing researches did , which leads to increment in the volume of model parameters and ignores the interactions among the blocks, 
 
 \begin{equation}
 \begin{split}
\mathcal{A} &\approx \mathcal{G} \times_{1} U_{1} \times_{2} U_{2} \times_{3} U_{3} \\&=\sum_{p=1}^{P} \sum_{q=1}^{Q} \sum_{r=1}^{R} g_{p q r} u_{1_{p}} \circ u_{2_{q}} \circ u_{3_{r}} \\&=\llbracket {\mathcal{G} ; U_{1}, U_{2}, U_{3}} \rrbracket
 \end{split}
\end{equation}

where, $ \mathcal{G} \times_{i} U_{i}$ denotes the  \textit{n-mode product} of $ \mathcal{G}\in R^{ P \times  Q \times R}$ with $U_{i}\in R^{ L_{i} \times  P }$, $L_{i}$ is the dimension of the \textit{i-th} mode of $A$; P, Q and R are the number of components (i.e., columns) in the factor matrices;  $\circ$  represents the vector outer product and $u_{i_{p}}$ is the \textit{p-th} column vector of $U_{i}$. Finally, the core tensor $ \mathcal{G}$, which can capture the hidden information and interaction characteristics among each mode of $\mathcal{A}$, is used to reconstruct the fusion graph as ${A}^{\prime}$ off-line though an unfolding-like operation, as shown in Fig.~\ref{fig:2l}(c). Specifically, in the unfolding-like operation, every three slices $ \mathcal{G}(:,:,m:m+2)$ are arranged in the $\textit{m-th}$ row of the new block matrix ${A}^{\prime}$. The objective function of Tucker decomposition for $\mathcal{A}$  is compactly formulated as:

\begin{equation}
\max _{\left\{\mathbf{U}_n \in \mathbb{S}\left(D_n, d_n\right)\right\}_{n \in[N]}}\left\|\mathcal{\mathcal { \mathcal { A }}} \times_{n \in[N]} \mathbf{U}_n^{\top}\right\|_F^2
\label{equation:4}
\end{equation}

where $\mathbb{S}(D, d)=\left\{\mathbf{U} \in \mathbb{R}^{D \times d} ; \quad \mathbf{U}^{\top} \mathbf{U}=\mathbf{I}_d\right\}$ is the Stiefel
manifold containing all rank-d orthonormal bases in $\mathbb{R}^{D}$ and $\|\cdot\|_F^2$ denotes the L2 (or Frobenius) norm returning the summation of the squared entries of its tensor argument. If $\left\{\mathbf{U}_n^{\mathrm{tckr}}\right\}_{n \in[N]}$ is solution to \ref{equation:4}, then the core tensor is expressed by:

\begin{equation}
\mathcal{G}^{\text {tckr }}:=\mathcal{\mathcal { A }}  \times_{n \in[N]} \mathbf{U}_n^{\mathrm{tckr}^{\top}}
\end{equation}

Specifically, the L1-Tucker decomposition \cite{chachlakis2019l1}, an improved Tucker decomposition model which replaces the L2 norm in the objective function of Tucker decomposition into L1 norm and has shown high efficiency, is adopted in this paper. The objective function of the L1-Tucker decomposition for  $\mathcal { A }$ is as follows:  %Fig.~\ref{fig:3l} depicts the differences between Tucker decomposition and L1-Tucker decomposition of a three-order tensor $\mathcal{X} \in R^{I \times J \times K}$. 

\begin{equation}
\max _{\left\{\mathbf{U}_n \in \mathbb{S}\left(D_n, d_n\right)\right\}_{n \in[N]}}\left\|\mathcal{\mathcal { A }} \times_{n \in[N]} \mathbf{U}_n^{\top}\right\|_1
\end{equation}

It is notable that, to keep the size of the graph unchangeable, during the tensor decomposition, the size of the core tensor is fixed as the as same as the original tensor.

%\begin{figure*}[!ht]
 % \centering
%\includegraphics[width=0.6\textwidth]{Fusion1}
 % \caption{(a) is a three-order tensor $\mathcal{X} \in R^{I \times J \times K}$. (b) is the Tucker decomposition. $A \in R^{I \times P}$, $B \in R^{J \times Q}$ and $C \in R^{K \times R}$ are the factor matrices, and the core tensor $ \mathcal{G}$  represents the relationship between these components. (c) is the L1-Tucker decomposition.}\label{fig:3l}
%\end{figure*}

\subsection{Spatial-Temporal Tensor Graph Convolutional Network}\label{}
\begin{figure*}[!ht]
  \centering
  \includegraphics[width=1\textwidth]{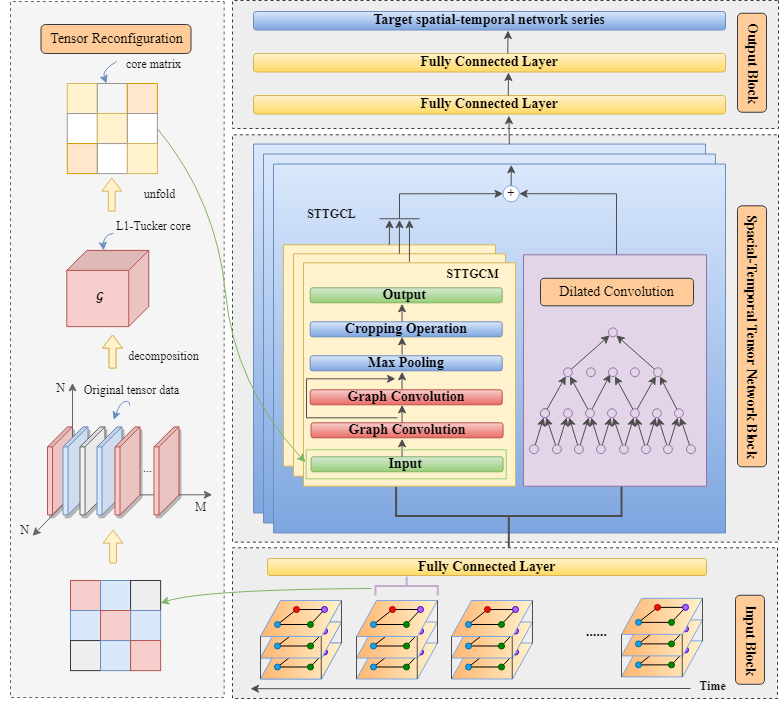}
  \caption{{Spatial-Temporal Tensor Graph Convolutional Network. The network consists of three blocks, from bottom to top, which are the input block, the spatial-temporal tensor network block and the output block, separately. The input block constructs the adjacent time steps into a local spatial-temporal matrix, and uses the core tensor attained through the tensor decomposition to reconstruct it. The reconstructed local spatial-temporal matrix is used as the input of STTGCM in the spatial-temporal tensor network block, and is combined with the dilated convolution module to enter the output block. Two fully connected layers are used to assemble the features extracted in the previous layers and provide the final outputs.}}\label{fig:4l}
\end{figure*}

 The Spatial-Temporal Synchronous Graph Convolutional Module (STSGCM) \cite{song2020spatial} is deployed in this paper to capture the localized spatial-temporal heterogeneous correlations in the graph reconstructed in Section 4.1. Furthermore, to learn the global spatial-temporal homogeneous correlations, we assemble a Dilated Convolution Module \cite{yu2017spatio} with STSGCM into a spatial- temporal tensor graph convolution layer (STTGCL). Fig~\ref{fig:4l} shows the specific structure of the model.

\subsubsection{Spatial-Temporal Tensor Graph Convolution Module}\label{}
Since the spatial-temporal graph constructed in Section 4.1 puts the nodes
at different time steps into a same environment without distinguishing
them, it obscures the time attribute of each node. To solve this problem, we equip position embedding \cite{gehring2017convolutional} to the spatial-temporal network series as Song \cite{song2020spatial} did, so that the model can take the spatial and temporal information into account, which can enhance the ability to model the spatial-temporal correlations. In STTGCM, GLU is selected as the activation function of graph convolution, so the graph convolution operation can be described as follows:

\begin{equation}
\begin{split}
% \begin{aligned}
&\operatorname{GCN}\left(h^{(l)}\right)=h^{(l+1)}\\&=\left(A^{\prime} h^{(l)} W_{1}+b_{1}\right) \otimes \sigma\left(A^{\prime} h^{(l)} W_{2}+b_{2}\right) \in R^{3 N \times C^{\prime}}
% \end{aligned}
\end{split}
\end{equation}
$A^{\prime} \in R^{ 3 N \times  3 N}$ is the graph adjacency matrix constructed in Section 4.1. $h^{l} \in R^{3 N \times C}$ belongs to the \textit{l-th} input of the graph convolution layer. $W_{1} \in R^{C \times C^{\prime}}$, $W_{2} \in R^{C \times C^{\prime}}$, $b_{1} \in R^{C^{\prime}}$, $b_{2} \in R^{C^{\prime}}$ are the weight parameters and bias parameters that can be obtained by learning. $\sigma$ represents the activation function and $\otimes$ represents the element product. Gated linear units control which nodes' information can be passed to the next layer.

Then, we can aggregate the complex spatial relationships by stacking $L$ layers. Here, we select the maximum pool as the aggregation operation. Finally, we cut the information of the previous and next time steps, because the information of these steps has been aggregated during the graph convolution operation, which would cause information redundancy if not deleted. The cutting form is shown in Fig~\ref{fig:1l}. There are T time steps, and every three neighbouring time steps are in a group. Then the outputs of STTGCM for the T-2 time step  are the final outputs. These outputs are connected in series into a matrix as the output of STTGCL layer. That can be formulated as:

\begin{equation}
Y=\left[Y_1, Y_2, \ldots, Y_{T-2}\right] \in \mathbb{R}^{(T-2) \times N \times C^{\prime}}
\end{equation}
where $Y_i \in \mathbb{R}^{N \times C^{\prime}}$ represents the outputs of the $i$-th STTGCM.

\subsubsection{Dilated Convolution Module}\label{}
\begin{figure}[!ht]
  \centering
  \includegraphics[width=0.4\textwidth]{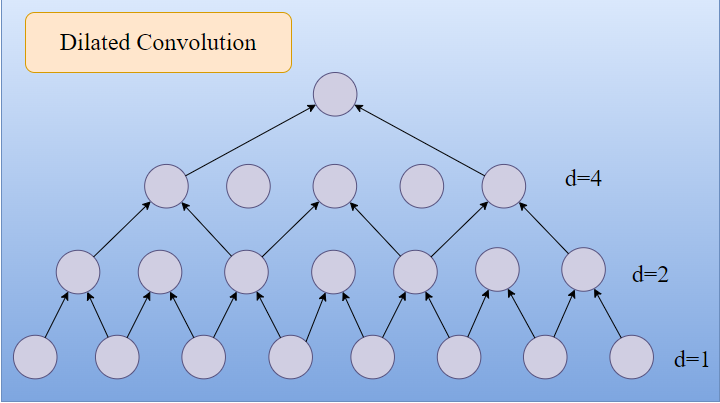}
  \caption{{Dilated Convolution.}}\label{fig:5l}
\end{figure}

STTGCM can  capture the heterogeneity in the local spatial-temporal graph effectively through multiple modules in different time periods, but ignored the global spatial-temporal homogeneity which also has a great influence in the prediction accuracy \cite{li2021spatial}. Hence, we introduce the dilated convolution, like many works \cite{yu2017spatio,wu2019graph},  to capture the global spatial-temporal homogeneity, as shown in Fig~\ref{fig:5l}. Dilated convolution allows the input of convolution to be sampled at intervals, and the dilated rate is controlled by $d$.

\subsubsection{Loss Function}\label{}
Huber Loss is a parametric Loss function used in regression problems, it has the advantage of enhancing the robustness of mean square error (MSE) to outliers. When the prediction deviation is less than $\delta$, it uses the squared error. When the prediction deviation is greater than $\delta$, the linear error is adopted. Hence, we choose Huber loss as the loss function.

\begin{center}
\begin{equation}
L\left(Y, Y^{\prime}\right)=\left\{\begin{array}{cl}
\frac{\left(Y^{\prime}-Y\right)^{2}}{2} & ,\left|Y^{\prime}-Y\right| \leq \delta \\
\delta\left|Y^{\prime}-Y\right|-\frac{\delta^{2}}{2} & ,\left|Y^{\prime}-Y\right|>\delta
\end{array}\right.
\end{equation}
\end{center}

$Y$ represents the basic truth value, $Y^{\prime}$ represents the prediction of the model, $\delta$ and is the threshold parameter that controls the square error loss range.

\section{Experiments}\label{}
In this section, we conducted a series of experiments on four real-world datasets, and evaluated the performances of the proposed model.
\subsection{Datasets}\label{}
% We verified the effectiveness of our proposed model on four highway traffic datasets \href{http://pems.dot.ca.gov/}{(http://pems.dot.ca.gov/)}. These data are collected from the Caltrans Performance Measurement System (PeMS).  The flow data is aggregated to 5 minutes, which means there are 12 points in the traffic flow data for one hour and 288 points for one day. Table~\ref{tab:datasets} provides the detailed information of the four datasets. One hour 12 continuous time steps historical data is used to predict next hour’s 12 continuous time steps data.
We verified the effectiveness of our proposed model on four highway traffic datasets (http://pems.dot.ca.gov/). These data are collected from the Caltrans Performance Measurement System (PeMS).  The flow data is aggregated to 5 minutes, which means there are 12 points in the traffic flow data for one hour and 288 points for one day. Table~\ref{tab:datasets} provides the detailed information of the four datasets. One hour 12 continuous time steps historical data is used to predict next hour’s 12 continuous time steps data.

\begin{table}[!ht]
\caption{Datasets}\label{tab:datasets}
\centering
\begin{tabular*}{\hsize}{@{\extracolsep{\fill}}|c |c |c |c|}
% \begin{tabular*}{\tblwidth}{lclclcl}
\hline
Datasets & Number of sensors & Time Steps & Time range\\ % Table header row
\hline
 PEMS03  & 358 & 26208 & 2018/9/1-2018/11/30 \\
\hline
 PEMS04  & 307 & 16992 & 2018/1/1-2018/2/28 \\
 \hline
 PEMS07  & 883 & 28224 & 2017/5/1-2017/8/31 \\
 \hline
 PEMS08  & 170 & 17856 & 2016/7/1-2016/8/31 \\
\hline
\end{tabular*}
\end{table}

\subsection{Models for Comparison}\label{}
Our proposed model will be compared with the following models:
\begin{itemize}[]
\item SVR \cite{chang2011libsvm}: Support Vector Regression is a classical traditional time series analysis model.
\item LSTM \cite{hochreiter1997long}: Long-Short Term Memory belongs to a typical kind of Recurrent Neural Networks.
\item DCRNN \cite{li2017diffusion}: Diffusion Convolution Recurrent Neural Network is a model that uses diffusion recurrent neural networks to capture spatial-temporal dependencies.
\item STGCN \cite{yu2017spatio}: Spatial-temporal Convolution Network integrates Chebnet and 2D convolution to obtain spatial-temporal correlations.
\item ASTGCN(r) \cite{guo2019attention}: Attention Based Spatial Temporal Graph
Convolutional Networks introduces spatial-temporal attention mechanisms into the graph convolution network model. Only recent
components of modeling periodicity is taken to keep fair
comparison
\item Graph WaveNet \cite{wu2019graph}: A framework combines
adaptive adjacency matrix into graph convolution
with 1D dilated convolution.
\item STSGCN \cite{wu2019graph}: Spatial-temporal Synchronous Graph Convolution Network uses local spatiotemporal sub-graph modules to independently model local correlations.
\item STFGNN \cite{li2021spatial}: Spatial-Temporal Fusion Graph Neural Networks integrates the fusion graph module and a new gate convolution module into a unified layer, which can learn the temporal and spatial dependencies to deal with long sequences.
\end{itemize}

\subsection{Experiment Settings}\label{}
To keep fair comparison with the previous works, all the data is split with the ratio 6:2:2 into  training sets, verification sets and test sets. All experiments are repeated on all datasets for ten times.

The number of time steps in a local temporal-spational network is selected as 3 and the model consists of 4 STTGCLs. Each STTGCM includes 3 graph convolutional operations with 64, 64, 64 filters respectively. The dilated coefficient of the dilated convolution module is 2. In the training process, we set the batch size to be 32, learning rate to be 0.003 and epoch equals to 5000, and we used an early-stop strategy. The threshold parameter of loss function $\delta$ is 1. All experiments were trained and tested under the computational environment with one Intel(R) Core(TM) i9-12900KF CPU @ 5.20Ghz and NVIDIA GeForce RTX 3090 Ti. Besides, in this work, we deploy the core tensor obtained by the tensor decomposition to reconstruct the graph adjacency matrix. Since the diagonal entries of the core tensor represent the connection of a node with itself at the same time step, which should be the largest theoretically. So we adjust them to be the max value 1 manually during the decomposition. 

The mean absolute error(MAE), the mean absolute percentage error
(MAPE) and the root mean square error (RMSE) are used to evaluate the prediction performances of the proposed model:
\begin{equation}
MAE=\frac{1}{n} \sum_{i=1}^{n}\left|Y-Y^{\prime}\right|
\end{equation}

\begin{equation}
MAPE=\frac{100\%}{n} \sum_{i=1}^{n}\left|\frac{Y-Y^{\prime}}{Y}\right|
\end{equation}

\begin{equation}
RMSE=\sqrt{\frac{1}{n} \sum_{i=1}^{n}\left(Y-Y^{\prime}\right)^{2}}
\end{equation}

where n is the total number of 5-min traffic flow data for prediction in test datasets. $Y$ and $Y^{\prime}$ are the true and predicted values respectively.

\subsection{Experimental Results ans Analysis}\label{}
\subsubsection{Prediction accuracy comparisons}\label{}
Table~\ref{tab:performance} shows the  prediction results comparison of different methods. It is obvious that our proposed model is always superior to others on all the four datasets.

\begin{table*}[h!]
\centering
\scriptsize
\caption{Performance Comparison of Baseline Model on four datasets.}\label{tab:performance}
\setlength{\tabcolsep}{2.5mm}{
\begin{tabular}{ |c | c|c|c|c|c|c|c|c|c|c| }
\hline
\multicolumn{2}{|c|}{Baseline methonds} & \multirow{2}{*}{SVR} & \multirow{2}{*}{LSTM} & \multirow{2}{*}{DCRNN} & \multirow{2}{*}{STGCN} & \multirow{2}{*}{ASTGCN} & \multirow{2}{*}{Graph WaveNet} & \multirow{2}{*}{STSGCN} & \multirow{2}{*}{STFGNN} & \multirow{2}{*}{STTGCN} \\% Table header row
  Datasets                  & Metrics   &                      &                        &                                &                         &                         &                               &                         &                         &                    \\            
\hline

\multirow{3}{*}{PEMS03}   & MAE & 21.07±0.00 & 21.33±0.24 & 18.18±0.15 & 17.49±0.46 & 17.69±1.43 & 19.85±0.03 & 17.48±0.15 & 16.77±0.09 & \textbf{15.20±0.17} \\
                          & MAPE(\%) & 21.51±0.46 & 23.33±4.23 & 18.91±0.82 & 17.15±0.45 & 19.40±2.24 & 19.31±0.49 & 16.78±0.20 & 16.30±0.09 & \textbf{14.95±0.56} \\
                          & RMSE & 35.29±0.02 & 35.11±0.50 & 30.31±0.25 & 30.12±0.70 & 29.66±1.68 & 32.94±0.18 & 29.21±0.56 & 28.34±0.46 & \textbf{26.60±0.58} \\
\hline
\multirow{3}{*}{PEMS04}   & MAE & 28.70±0.01 & 27.14±0.20 & 24.70±0.22 & 22.70±0.64 & 22.93±1.29 & 25.45±0.03 & 21.19±0.10 & 19.83±0.06 & \textbf{19.08±0.20} \\
                          & MAPE(\%) & 19.20±0.01 & 18.20±0.40 & 17.12±0.37 & 14.59±0.21 & 16.56±1.36 & 17.29±0.24 & 13.90±0.05 & 13.02±0.05 & \textbf{12.61±0.11} \\
                          & RMSE & 44.56±0.01 & 41.59±0.21 & 38.12±0.26 & 35.55±0.75 & 35.22±1.90 & 39.70±0.04 & 33.65±0.20 & 31.88±0.14 & \textbf{30.96±0.41} \\
\hline                          
\multirow{3}{*}{PEMS07}   & MAE & 32.49±0.00 & 29.98±0.42 & 25.30±0.52 & 25.38±0.49 & 28.05±2.34 & 26.85±0.05 & 24.26±0.14 & 22.07±0.11 & \textbf{20.36±0.13} \\
                          & MAPE(\%) & 14.26±0.03 & 13.20±0.53 & 11.66±0.33 & 11.08±0.18 & 13.92±1.65 & 12.12±0.41 & 10.21±1.65 & 9.21±0.07 & \textbf{8.62±0.10} \\
                          & RMSE & 50.22±0.01 & 45.84±0.57 & 38.58±0.70 & 38.78±0.58 & 42.57±3.31 & 42.78±0.07 & 39.03±0.27 & 35.80±0.18 & \textbf{33.92±0.23} \\
\hline                          
\multirow{3}{*}{PEMS08}   & MAE & 23.25±0.01 & 22.20±0.18 & 17.86±0.03 & 18.02±0.14 & 18.61±0.40 & 19.13±0.08 & 17.13±0.09 & 16.64±0.09 & \textbf{15.09±0.09} \\
                          & MAPE(\%) & 14.64±0.11 & 14.20±0.59 & 11.45±0.03 & 11.40±0.10 & 13.08±1.00 & 12.68±0.57 & 10.96±0.07 & 10.60±0.06 & \textbf{9.77±0.14} \\
                          & RMSE & 36.16±0.02 & 34.06±0.32 & 27.83±0.05 & 27.83±0.20 & 28.16±0.48 & 31.05±0.07 & 26.80±0.18 & 26.22±0.15 & \textbf{24.37±0.16} \\
\hline
\end{tabular}}
\end{table*}

 The proposed STTGCN model works best on the PEMS08 datasets, which outperforms the second performer STFGNN with improvements of 1.55 and 1.85 on MAE and RMSE. 
 The STTGCN achieves improved average performances on MAE, MAPE, and RMSE,compared to other methods, which outperforms the second performer STFGNN with improvements of 1.395, 0.795, and 1.598 on MAE, MAPE and RMSE, and outperforms the STSGCN with improvement of 2.43 on  RMSE. The traditional forecasting methods (SVR and LSTM) that only considered the temporal correlations while ignored the spatial correlations perform the worst. The experimental results confirm that our approach can construct more informative graph for the network and learn richer spatial-temporal correlations of traffic flow data to provide more accurate forecasting. STSGCN and STFGNN achieve better performances than DCRNN, STGCN, ASTGCN and Graph WaveNet, because they can capture the spatial-temporal correlations directly through a fusion graph model instead of using different types of neural network modules and sophisticated mechanisms for modeling spatial and temporal correlations separately. STFGNN outperforms STSGCN mainly because that it assembled a gated convolution to learn the global homogeneous correlations except for the localized spatial-temporal dependencies learned by STSGCN. 

\begin{figure*}[!ht]
  \centering
  \includegraphics[width=1.09\textwidth]{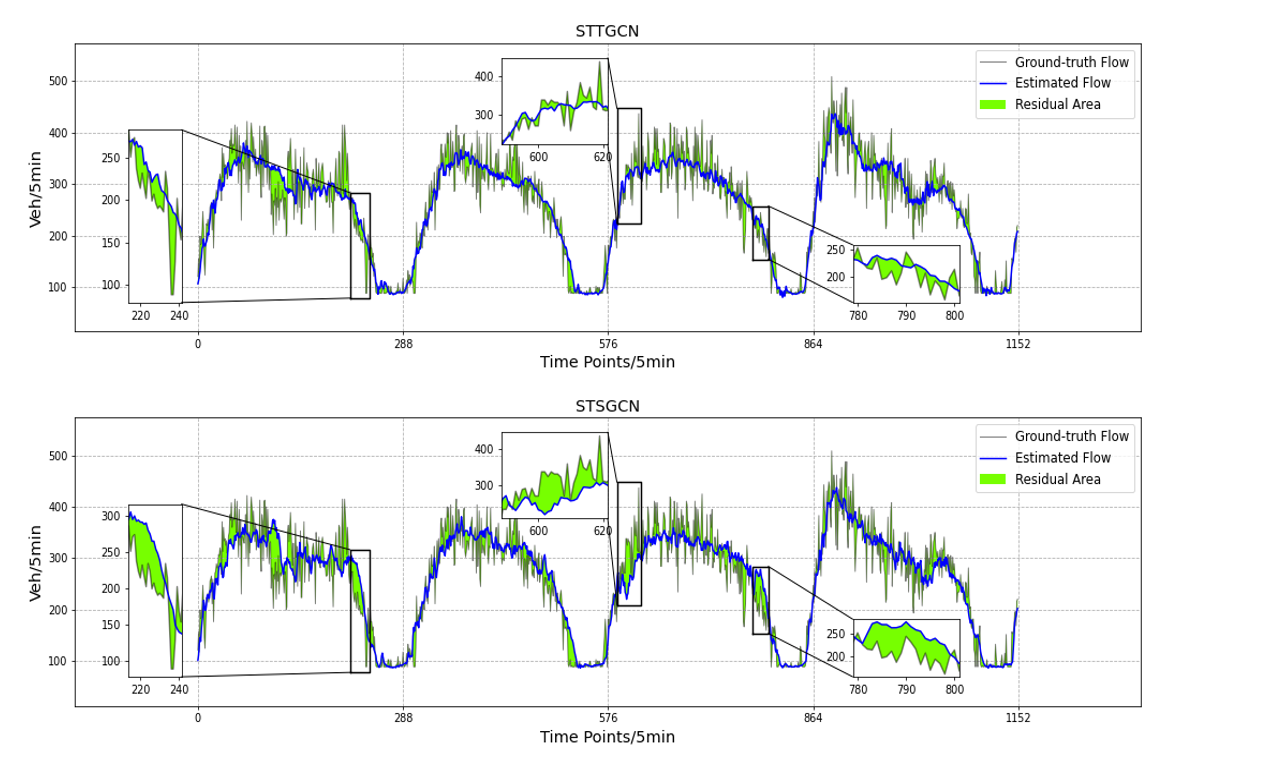}
  \caption{{The prediction details of (a) STTGCN and (b) STSGCN for a certain road segment on PEMS04. The green area represents the residual between the ground-truth traffic flow data and the predicted ones.}}\label{fig:6l}
\end{figure*}

 Fig~\ref{fig:6l} shows the   prediction details of STTGCN and STSGCN for a certain road segment on PEMS04. It is interesting to find that STTGCN outperforms STSGCN prominently in the morning and evening peaks. This means our method may be more robust when dealing with the large-scale traffic flow.

\subsubsection{Comparison of the model parameters quantities}\label{}
Table~\ref{tab:parameters} presents the  model parameters quantities of STTGCN, STFGNN and STSGCN. It can be seen that the necessary model parameters of  STTGCN are much less than the other two methods especially for the large-scale dataset like PEMS07, which means STTGCN is superior in the computational cost for higher training efficiency. This is mainly because that STSGCN and STFGNN involves the mask matrix learning process which produces too many parameters. STTGCN deploys a tensor decomposition operation to reconstruct and update the graph adjacency matrix offline, which helps to avoid such a learning process and hence reduce the model computational cost as increasing the prediction accuracy.

\begin{table}[!ht]
\caption{Model training parameters quantities.}\label{tab:parameters}
\begin{tabular*}{\hsize}{@{\extracolsep{\fill}}|c|c|c|c|c|}
% \begin{tabular*}{\tblwidth}{ccccc}
\hline
\diagbox{Models}{datasets} & PEMS03 & PEMS04 & PEMS07 & PEMS08 \\
\hline
STSGCN & 3496212 & 2872686 & 15358062 & 1661332\\
\hline
STFGNN & 4968652 & 3873580 & 25918252 & 1756108\\
\hline
STTGCN & \underline{\textbf{1255308}} & \underline{\textbf{1242252}} & \underline{\textbf{1389708}} & \underline{\textbf{1207108}} \\
\hline
\end{tabular*}
\end{table}

\begin{figure*}[!ht]
	\centering
		\includegraphics[scale=0.95]{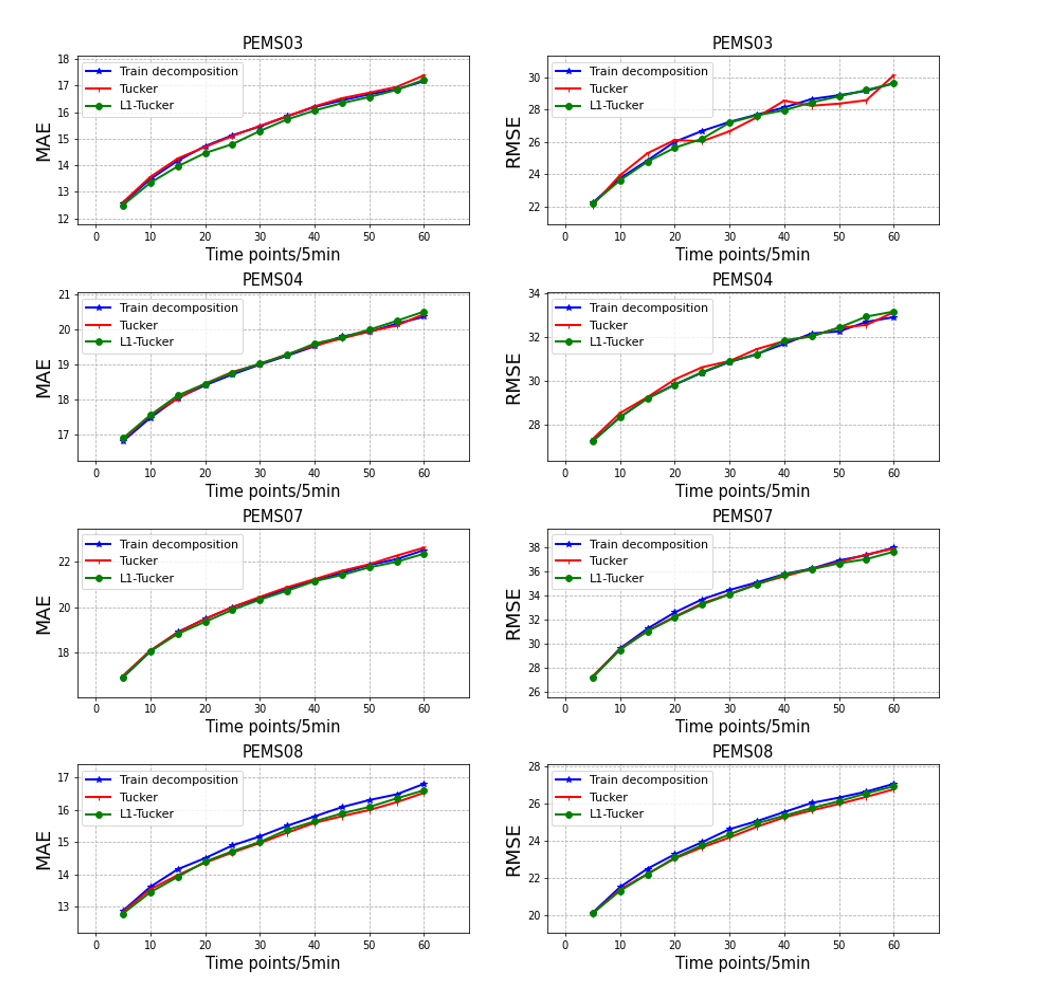}
	  \caption{Performance comparison of different core tensors}\label{fig:7l}
\end{figure*}

\subsubsection{Comparison of adopting different tensor decomposition models}\label{}
Adopting the tensor decomposition to reconstruct the adjacency graph is a main innovation point and highlight of this work. Hence, choosing a proper decomposition model is of great significance. Fig~\ref{fig:7l} presents the performances of using the following three different tensor factorization models on the four datasets.

\begin{itemize}[]
\item Tucker \cite{kolda2009tensor}: a traditional tensor decomposition method.
\item  L1-Tucker \cite{chachlakis2019l1}: the L1 norm constrained Tucker decomposition method.
\item Train decomposition \cite{bigoni2016spectral}: the original high-dimensional tensor is decomposed into the product of multiple three-dimensional tensors.
\end{itemize}

We tested the above three tensor decomposition methods for graph reconstructing on predicting a random one-hour period of traffic flow on all the four datasets. Fig~\ref{fig:7l} shows the MAE and RMSE in each time point. In the PEMS08 dataset, we can see that the Train decomposition based method, whose prediction error rises rapidly after the second time points, is inferior to the Tucker decomposition based methods. This may be because that the Train decomposition is mainly suitable for processing super high-order($\ge4$) tensors but not the 3-order tensor constructed in this paper. The Tucker and the L1-Tucker methods show little differences in the performance of prediction accuracy on all the datasets, but L1-Tucker trends to be more robust in the PEMS03 and PEMS04 datasets, especially in PEMS03.

\begin{figure}[htbp]
	\centering
		\includegraphics[scale=0.4]{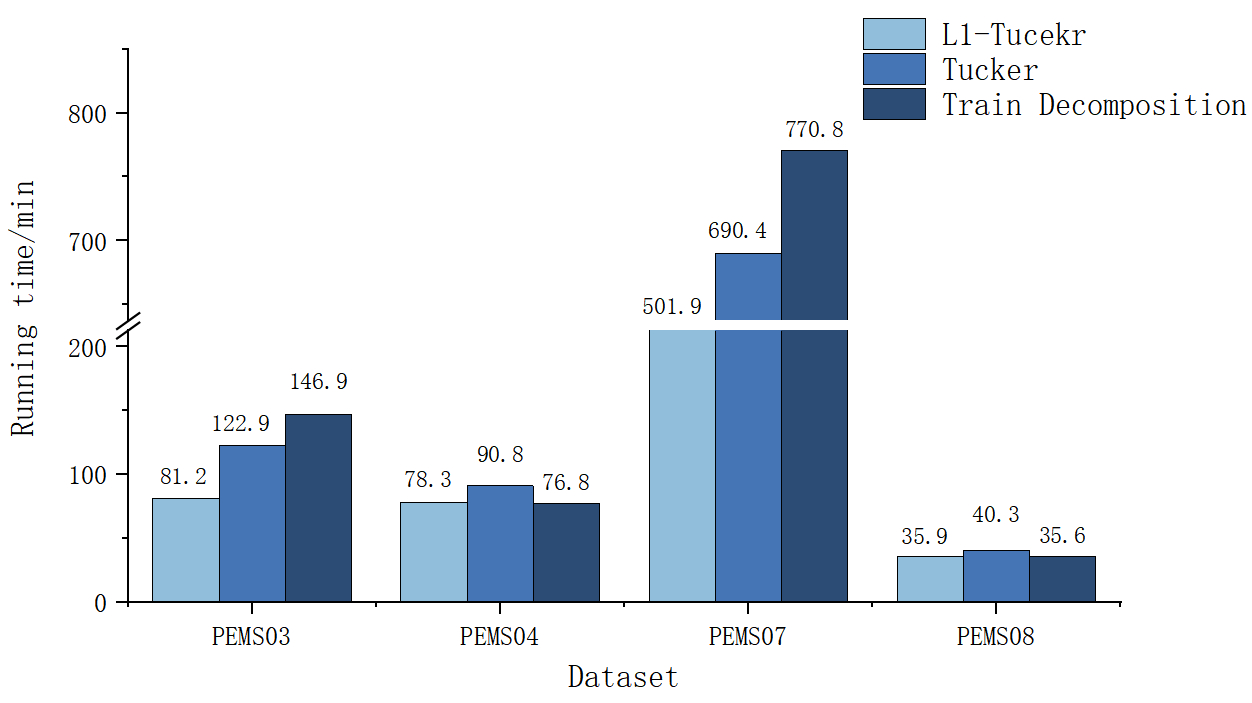}
	  \caption{Running times for different decomosition methods in each datasets}\label{fig:8l}
\end{figure}

For further comparison, we tested the computational consumption  of the three models. As shown in Fig~\ref{fig:8l}, to achieve the same prediction accuracy, L1-Tucker is of the highest efficiency in the large-scale PEMS03 and PEMS07 datasets. Hence, leveraging the effectiveness and efficiency for prediction, L1-Tucker is deployed in this paper to reconstruct and improve the graph model.

\subsubsection{The node relationships in the reconstructed graph}\label{}
\begin{figure}[htbp]
	\centering
		\includegraphics[scale=0.35]{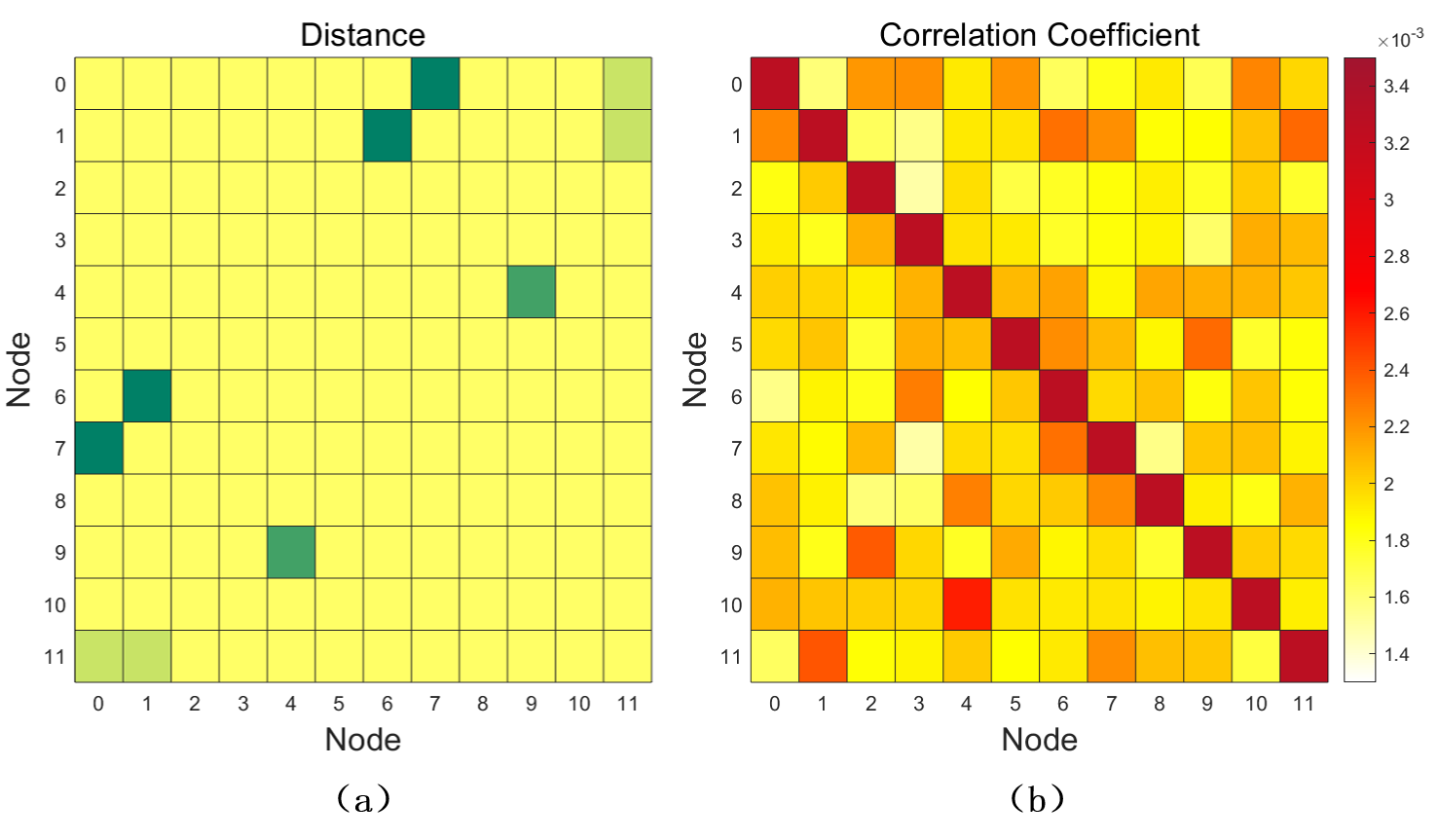}
	  \caption{The node relationships in the physical space and reconstructed adjacency graph:(a) the connectivities and corresponding distances among detectors in the physical space;(b) the relationships among detectors in the reconstructed graph.}\label{fig:9l}
\end{figure}

In order to investigate the role of tensor decomposition in our model intuitively, we shown a hotmap (Fig~\ref{fig:9l}) which presents the relationships of 12 random detectors  in the reconstructed adjacency graph for the PEMS08 dataset. Fig~\ref{fig:9l}(a) shows the real-world physical distances between neighboring detectors, which is used to construct the given spatial graph. Darker colors indicate farther distances, except for that the yellow grids, whose values equal to 0, represent there is no connection between the corresponding two nodes. Fig~\ref{fig:9l}(b) shows the reconstructed adjacency graph by our model, the i-\textit{th} row represents the correlation magnitudes between each detector and the i-\textit{th} detector. For example, the relationship between the 0\textit{th} and 7\textit{th}  detectors is weaker than that between the 1\textit{th} and 11\textit{th} detectors. This is mainly because the distances between the 0\textit{th} and 7\textit{th} detectors are much further as shown in Fig~\ref{fig:9l}(a). Besides, there are also relatively strong relationships demonstrated between some detectors that are not connective in the physical space (e.g. the 4\textit{th} and 10\textit{th} detectors) and the relationships between two closely connected detectors may be not always that strong(e.g. the 0\textit{th} and 11\textit{th} detectors). This means the reconstructed adjacency graph learns more informative relationships not only based on the physical connects but also considering some other factors such as the road capacity,traffic control, artificial preferences, etc.

\section{Conclusion}\label{}
In this paper, we proposed a new graph convolutional network model for spatial-temporal traffic flow forecasting. It can not only capture the underlying spatial-temporal dependencies effectively through an informative graph, but also learn the localized spatial-temporal heterogeneity and the global spatial-temporal homogeneity simultaneously. 

The main findings of this paper are: (1) the graph reconstructed by the core tensor obtained through tensor decomposition, especially the L1-Tucker decomposition, is much more informative than the original given spatial graph; (2) deploying the graph reconstructed and updated by tensor decomposition is beneficial for enhancing both the effectiveness and efficiency of traffic data forecasting; (3) learning the localized heterogeneous and global homogeneous spatial-temporal correlations simultaneously through an assembling model can improve traffic data forecasting accuracy, which is consistent with the conclusions of previous works. 

Above all, this paper suggests a new way of utilizing the tensor decomposition method to improve the graph convolutional network model for traffic data forecasting. However, the sub-graph information constituting the fusion graph is also important, so we will explore how to construct more informative sub-graphs according to the spatial and temporal properties of traffic data in the future (e.g. the functional similarity graph / the transportation connectivity graph). In addition, how to aggregate the spatial-temporal information is of great significance in the graph convolutional network model, hence we will continuously devote to investigating more efficient and effective learning frameworks.

\bibliographystyle{IEEEtran}
\bibliography{IEEEabrv}

\end{document}